\documentclass[journal]{IEEEtran}

\usepackage{ifpdf}

\ifCLASSINFOpdf
   \usepackage[pdftex]{graphicx}

\else

\fi

\usepackage[utf8]{inputenc} 
\usepackage[T1]{fontenc}    
\usepackage{hyperref}       
\usepackage{url}            
\usepackage{booktabs}       
\usepackage{amsfonts}       
\usepackage{nicefrac}       
\usepackage{microtype}      

\usepackage{amsmath}
\usepackage{amssymb}
\usepackage{amsthm}
\usepackage{bm}
\usepackage{mathtools}
\usepackage{dsfont}
\usepackage{multirow}
\usepackage{leftidx}
\usepackage{soul}

\usepackage{graphicx}
\usepackage{caption}
\usepackage{subcaption}

\usepackage{xargs}                      
\usepackage[pdftex,dvipsnames]{xcolor}  
\usepackage[colorinlistoftodos,prependcaption,textsize=tiny]{todonotes}
\newcommandx{\unsure}[2][1=]{\todo[linecolor=red,backgroundcolor=red!25,bordercolor=red,#1]{#2}}
\newcommandx{\change}[2][1=]{\todo[linecolor=blue,backgroundcolor=blue!25,bordercolor=blue,#1]{#2}}
\newcommandx{\info}[2][1=]{\todo[linecolor=OliveGreen,backgroundcolor=OliveGreen!25,bordercolor=OliveGreen,#1]{#2}}
\newcommandx{\improvement}[2][1=]{\todo[linecolor=Plum,backgroundcolor=Plum!25,bordercolor=Plum,#1]{#2}}
\newcommandx{\thiswillnotshow}[2][1=]{\todo[disable,#1]{#2}}

\providecommand{\mb}[1]{\mathbf{#1}}

\providecommand{\mbx}{\mb{x}}
\providecommand{\mby}{\mb{y}}
\providecommand{\mbz}{\mb{z}}

\usepackage{algorithmic}

\usepackage{multirow}
\usepackage{algorithm}
\newenvironment{sequation}{\begin{equation}\small}{\end{equation}}

\usepackage{array}

\usepackage{url}
\usepackage{capt-of}
\usepackage{wrapfig}
\usepackage{rotating}

\usepackage{multirow}

\makeatletter
\def\ps@IEEEtitlepagestyle{
  \def\@oddfoot{\mycopyrightnotice}
  \def\@evenfoot{}
}
\def\mycopyrightnotice{
  {\footnotesize
  \begin{minipage}{\textwidth}
  \centering
  Copyright~\copyright~2017 IEEE. Personal use of this material is permitted. However, permission to use this material for any other purposes must be obtained from the IEEE by sending a request to pubs-permissions@ieee.org.
  \end{minipage}
  }
}

\begin{document}

\title{Convolutional Recurrent Neural Networks for Dynamic MR Image Reconstruction}

%
%
%

\author{Chen~Qin*\textsuperscript{$\dagger$},
        Jo~Schlemper*, Jose Caballero, Anthony N. Price, Joseph V. Hajnal and Daniel Rueckert~\IEEEmembership{Fellow,~IEEE}
\thanks{This work was supported by EPSRC programme grant SmartHeart (EP/P001009/1). C. Qin is supported by China Scholarship Council (CSC).}
\thanks{\textsuperscript{$\dagger$}Corresponding author: Chen Qin. (Email address: c.qin15@imperial.ac.uk)}
\thanks{*These authors contributed equally to this work.}
\thanks{C. Qin, J. Schlemper, J. Caballero and D. Rueckert are with the Biomedical Image Analysis Group, Department of Computing, Imperial College London, SW7 2AZ London, UK.} \thanks{J. V. Hajnal, and A. N. Price are with with the Division of Imaging Sciences and Biomedical Engineering Department, King’s College London, St. Thomas’ Hospital, SE1 7EH London, U.K.}}

%



\maketitle

\begin{abstract}

Accelerating the data acquisition of dynamic magnetic resonance imaging (MRI) leads to a challenging ill-posed inverse problem, which has received great interest from both the signal processing and machine learning communities over the last decades. The key ingredient to the problem is how to exploit the temporal correlations of the MR sequence to resolve aliasing artefacts. Traditionally, such observation led to a formulation of an optimisation problem, which was solved using iterative algorithms. Recently, however, deep learning based-approaches have gained significant popularity due to their ability to solve general inverse problems. In this work, we propose a unique, novel convolutional recurrent neural network (CRNN) architecture which reconstructs high quality cardiac MR images from highly undersampled k-space data by jointly exploiting the dependencies of the temporal sequences as well as the iterative nature of the traditional optimisation algorithms. In particular, the proposed architecture embeds the structure of the traditional iterative algorithms, efficiently modelling the recurrence of the iterative reconstruction stages by using recurrent hidden connections over such iterations. In addition, spatio-temporal dependencies are simultaneously learnt by exploiting bidirectional recurrent hidden connections across time sequences. The proposed method is able to learn both the temporal dependency and the iterative reconstruction process effectively with only a very small number of parameters, while outperforming current MR reconstruction methods in terms of reconstruction accuracy and speed.

\end{abstract}


%
\IEEEpeerreviewmaketitle

\section{Introduction}

\IEEEPARstart{M}{agnetic} Resonance Imaging (MRI) is a non-invasive imaging technique which offers excellent spatial resolution and soft tissue contrast and is widely used for clinical diagnosis and research. Dynamic MRI attempts to reveal both spatial and temporal profiles of the underlying anatomy, which has a variety of applications such as cardiovascular imaging and perfusion imaging. However, the acquisition speed is fundamentally limited due to both hardware and physiological constraints as well as the requirement to satisfy the Nyquist sampling rate. Long acquisition times are not only a burden for patients but also make MRI susceptible to motion artefacts. 

In order to accelerate MRI acquisition, most approaches consider undersampling the data in $k$-space (frequency domain). Due to the violation of the Nyquist sampling theorem, undersampling introduces aliasing artefacts in the  image domain. Images can be subsequently reconstructed by solving an optimisation problem that regularises the solution with assumptions on the underlying data, such as smoothness, sparsity or, for the case of dynamic imaging, spatio-temporal redundancy. Past literature has shown that exploiting spatio-temporal redundancy can greatly improve image reconstruction quality compared to compressed sensing (CS) based single frame reconstruction methods \cite{jung2007improved, lingala2011ktslr}. {However, the challenges of these optimisation based approaches are the following: firstly, the regularisation functions and their hyper-parameters must be carefully selected, which are problem-specific and non-trivial. For example, over-imposing sparsity or $\ell_1$ penalties can lead to cartoon-like/staircase artefacts.} 
Secondly, the reconstruction speeds of these methods are often slow due to requirement to solve iterative algorithms. Proposing a robust iterative algorithm is still an active area of research. 

In comparison, deep learning methods are gaining popularity for their accuracy and efficiency. Unlike traditional approaches, the prior information and regularisation are learnt implicitly from data, {without having to specify them in the training objective.}
However, so far only a handful of approaches exist \cite{schlemper2017dynamic,batenkov2017csc} for dynamic reconstruction. Hence, the applicability of deep learning models to this problem is yet to be fully explored. {In addition, many proposed deep learning architectures are often generic and are not optimised for specific applications.} In particular, a core question for dynamic reconstruction is how to optimally exploit spatio-temporal redundancy. {By designing a network architecture and regulating the mechanics of network layers to efficiently learn such spatio-temporal data representation, the network should gain a boost in performances.} 

In this work, we propose a novel convolutional recurrent neural network (CRNN) method to reconstruct high quality dynamic MR image sequences from undersampled data, termed \emph{CRNN-MRI}. Firstly, we formulate a general optimisation problem for solving accelerated dynamic MRI based on variable splitting and alternate minimisation. We then show how this algorithm can be seen as a network architecture. In particular, the proposed method consists of a CRNN block which acts as the proximal operator and a data consistency layer corresponding to the classical data fidelity term. In addition, the CRNN block employs recurrent connections across each iteration step, allowing reconstruction information to be shared across the multiple iterations of the process. Secondly, we incorporate bidirectional convolutional recurrent units evolving over time to exploit the temporal dependency of the dynamic sequences and effectively propagate the contextual information across time frames of the input. As a consequence, the unique CRNN architecture jointly learns representations in a recurrent fashion evolving over both \emph{time sequences} as well as \emph{iterations} of the reconstruction process, effectively combining the benefits of traditional iterative methods and deep learning. 

To the best of our knowledge, this is the first work applying RNNs for dynamic MRI reconstruction. The contributions of this work are the following: Firstly, we view the optimisation problem of dynamic data as a recurrent network and describe a novel CRNN architecture which simultaneously incorporates the {recurrence existing in both temporal and iteration sequential steps}. Secondly, we demonstrate that the proposed method shows promising results and improves upon the current state-of-the-art dynamic MR reconstruction methods both in reconstruction accuracy and speed. Finally, we compare our architecture to 3D CNN which does not impose the recurrent structure. We show that the proposed method outperforms the CNN at different undersampling rates and speed, while requiring significantly fewer parameters.

\section{Related Work}
One of the main challenges associated with recovering an uncorrupted image is that both the undersampling strategy and a-priori knowledge of appropriate properties of the image need to be taken into account. 
Methods like k-t BLAST and k-t SENSE \cite{tsao2003blast} take advantage of a-priori information about the x-f support obtained from the training data set in order to prune a reconstruction to optimally reduce aliasing. 
An alternative popular approach is to exploit temporal redundancy to unravel from the aliasing by using CS approaches \cite{jung2007improved, caballero2014dictionary} or CS combined with low-rank approaches \cite{lingala2011ktslr, otazo2015lpluss}. The class of methods which employ CS to the MRI reconstruction is termed as CS-MRI \cite{lustig2008compressed}. They assume that the image to be reconstructed has a sparse representation in a certain transform domain, and they need to balance sparsity in the transform domain against consistency with the acquired undersampled k-space data. For instance, an example of successful methods enforcing sparsity in x-f domain is k-t FOCUSS \cite{jung2007improved}. A low rank and sparse reconstruction scheme (k-t SLR) \cite{lingala2011ktslr} introduces non-convex spectral norms and uses a spatio-temporal total variation norm in recovering the dynamic signal matrix. Dictionary learning approaches were also proposed to train an over-complete basis of atoms to optimally sparsify spatio-temporal data \cite{caballero2014dictionary}. These methods offer great potential for accelerated imaging, however, they often impose strong assumptions on the underlying data, requiring nontrivial manual adjustments of hyperparameters depending on the application. In addition, it has been observed that these methods tend to result in blocky \cite{hammernik2017} and unnatural reconstructions, and their reconstruction speed is often slow. Furthermore, these methods are not able to exploit the prior knowledge that can be learnt from the vast number of MRI exams routinely performed, which should be helpful to further guide the reconstruction process.  

Recently, deep learning-based MR reconstruction has gained popularity due to its promising results for solving inverse and compressed sensing problems. In particular, two paradigms have emerged: the first class of approaches proposes to use convolutional neural networks (CNNs) to learn an end-to-end mapping, where architectures such as SRCNN {\cite{dong2014sr}} or U-net {\cite{ronneberger2015unet}} are often chosen for MR image reconstruction {\cite{lee2017deep,han2017deep,wang2016,wang2016c}}. The second class of approaches attempts to make each stage of iterative optimisation learnable by unrolling the end-to-end pipeline into a deep network \cite{hammernik2017, adler2017learned, NIPS2016_6406, schlemper2017deep, adler2017solving}.
For instance, Hammernik et al. \cite{hammernik2017} introduced a trainable formulation for accelerated parallel imaging (PI) based MRI reconstruction termed variational network, which embedded a CS concept within a deep learning approach. ADMM-Net \cite{NIPS2016_6406} was proposed by reformulating an alternating direction method of multipliers (ADMM) algorithm to a deep network, where each stage of the architecture corresponds to an iteration in the ADMM algorithm. More recently, Schlemper et al. \cite{schlemper2017deep} proposed a cascade network which simulated the iterative reconstruction of dictionary learning-based methods and were later extended for dynamic MR reconstructions \cite{schlemper2017dynamic}. Most approaches so far have focused on 2D images, whereas  only a few approaches exist for dynamic MR reconstruction \cite{schlemper2017dynamic,batenkov2017csc}. While they show promising results, the optimal architecture, training scheme and configuration spaces are yet to be fully explored. 

{More recently, several methods on 2D MR image reconstruction were proposed \cite{hammernik2017,aggarwal2017modl,mardani2017recurrent}, which share similar idea with our proposed method that integrates data fidelity term and regularisation term into a single deep network so that to enable the end-to-end training. In contrast to these methods which use shared parameters over iterations, as we will show, our architecture integrates hidden connections over optimisation iterations to propagate learnt representations across both iteration and time, whereas such information is discarded in the other methods.}
Such proposed architecture enables the information used for the reconstruction at each iteration to be shared across all stages of the reconstruction process, aiming for an iterative algorithm that can fully benefit from information extracted at all processing stages. As to the nature of the proposed RNN units, previous work involving RNNs only updated the hidden state of the recurrent connection with a fixed input \cite{gregor2015draw, liang2015recurrent,kuen2016recurrent}, while the proposed architecture progressively updates the input as the optimisation iteration increases. In addition, previous work only modelled the recurrence of iteration \emph{or} time \cite{huang2015bidirectional} exclusively, whereas the proposed method jointly exploits both dimensions, yielding a unique architecture suitable for the dynamic reconstruction problem. 

\section{Convolutional Recurrent Neural Network for MRI reconstruction}

\subsection{Problem Formulation}
Let ${\bf{x}} \in \mathds{C}^D$ denote a sequence of complex-valued MR images to be reconstructed, represented as a vector with $D=D_xD_yT$, and let ${\bf{y}} \in \mathds{C}^M$ $(M << D)$ represent the undersampled k-space measurements, where $D_x$ and $D_y$ are width and height of the frame respectively and $T$ stands for the number of frames. Our problem is to reconstruct $\bf{x}$ from $\bf{y}$, which is commonly formulated as an unconstrained optimisation problem of the form:
\begin{equation}
  \label{eq:sparse_coding}
\begin{aligned}
& \underset{\mbx}{\text{argmin}} & & \mathcal{R}(\mbx) + \lambda \| \mby - \mb{F}_u \mbx \|^2_2
\end{aligned}
\end{equation}
Here $\mb{F}_u$ is an undersampling Fourier encoding matrix, $\mathcal{R}$ expresses regularisation terms on $\mbx$ and $\lambda$ allows the adjustment of data fidelity based on the noise level of the acquired
measurements $\mby$. For CS and low-rank based approaches, the regularisation terms $\mathcal{R}$ often employed are $\ell_0$ or $\ell_1$ norms in the sparsifying domain of $\mbx$ as well as the rank or nuclear norm of $\mb{x}$ respectively. In general, Eq. \ref{eq:sparse_coding} is a non-convex function and hence, the variable splitting technique is usually adopted to decouple the fidelity term and the regularisation term. By introducing an auxiliary variable $\bf{z}$ that is constrained to be equal to $\bf{x}$, Eq. \ref{eq:sparse_coding} can be reformulated to minimize the following cost function via the penalty method:

\begin{equation}
\label{eq: penalty_function}
\underset{\mbx,\mbz}{\text{argmin }} \mathcal{R}(\mbz) + \lambda \| \mby - \mb{F}_u \mbx \|_2^2 + \mu \|\mbx-\mbz\|^2_2
\end{equation}
where $\mu$ is a penalty parameter. By applying alternate minimisation over $\bf{x}$ and $\bf{z}$, Eq. \ref{eq: penalty_function} can be solved via the following iterative procedures:
\begin{subequations}
\label{eq:alternate_minimisation} 
\begin{align}
\mbz^{(i+1)} & = \underset{\mbz}{\text{argmin }}   \mathcal{R}(\mbz) + \mu \| \mbx^{(i)} - \mbz \|^2_2  \label{eq:proximal_operator}
\\ 
\mbx^{(i+1)} & = \underset{\mbx}{\text{argmin }}  \lambda \| \mby - \mb{F}_u \mbx \|^2_2 + \mu \| \mbx - \mbz^{(i+1)} \|^2_2 \label{eq:data_fidelity}
\end{align}
\end{subequations}
where $\mbx^{(0)} = \mbx_u = {\bf{F}}_u^H y$ is the zero-filled reconstruction taken as an initialisation and $\mbz$ can be seen as an intermediate state of the optimisation process. For MRI reconstruction, Eq. \ref{eq:data_fidelity} is often regarded as a \emph{data consistency} (DC) step where we can obtain the following closed-form solution \cite{schlemper2017deep}:
\begin{equation}
\begin{array}{l}
\mbx^{(i+1)} = \textnormal{DC}(\mbz^{(i)};\mby, \lambda_0, \Omega) = \mb{F}^{H}\mb{\Lambda}\mb{F}\mbz^{(i)} + \frac{\lambda_0}{1 + \lambda_0} \mb{F}_u^{H}\mby, 
\\
\bm{\Lambda}_{kk} =
\begin{cases}
  1 & \text{if } k \not \in \Omega \\
  \frac{1}{1+\lambda_0} & \text{if } k \in \Omega
\end{cases}
\label{eq:dc_fnc}
\end{array}
\end{equation}
in which $\mb{F}$ is the full Fourier encoding matrix (a discrete Fourier transform in this case), $\lambda_0 = \lambda/\mu$ is a ratio of regularization parameters from Eq. \ref{eq:dc_fnc}, $\Omega$ is an index set of the acquired $k$-space samples and $\Lambda$ is a diagonal matrix. Please refer to  \cite{schlemper2017deep} for more details of formulating Eq. \ref{eq:dc_fnc} as a data consistency layer in a neural network. Eq. \ref{eq:proximal_operator} is the proximal operator of the prior $\mathcal{R}$, and instead of explicitly determining the form of the regularisation term, we propose to directly learn the proximal operator by using a convolutional recurrent neural network (CRNN). 

\begin{figure*}[!t]
\centering
\includegraphics[width=0.6\linewidth]{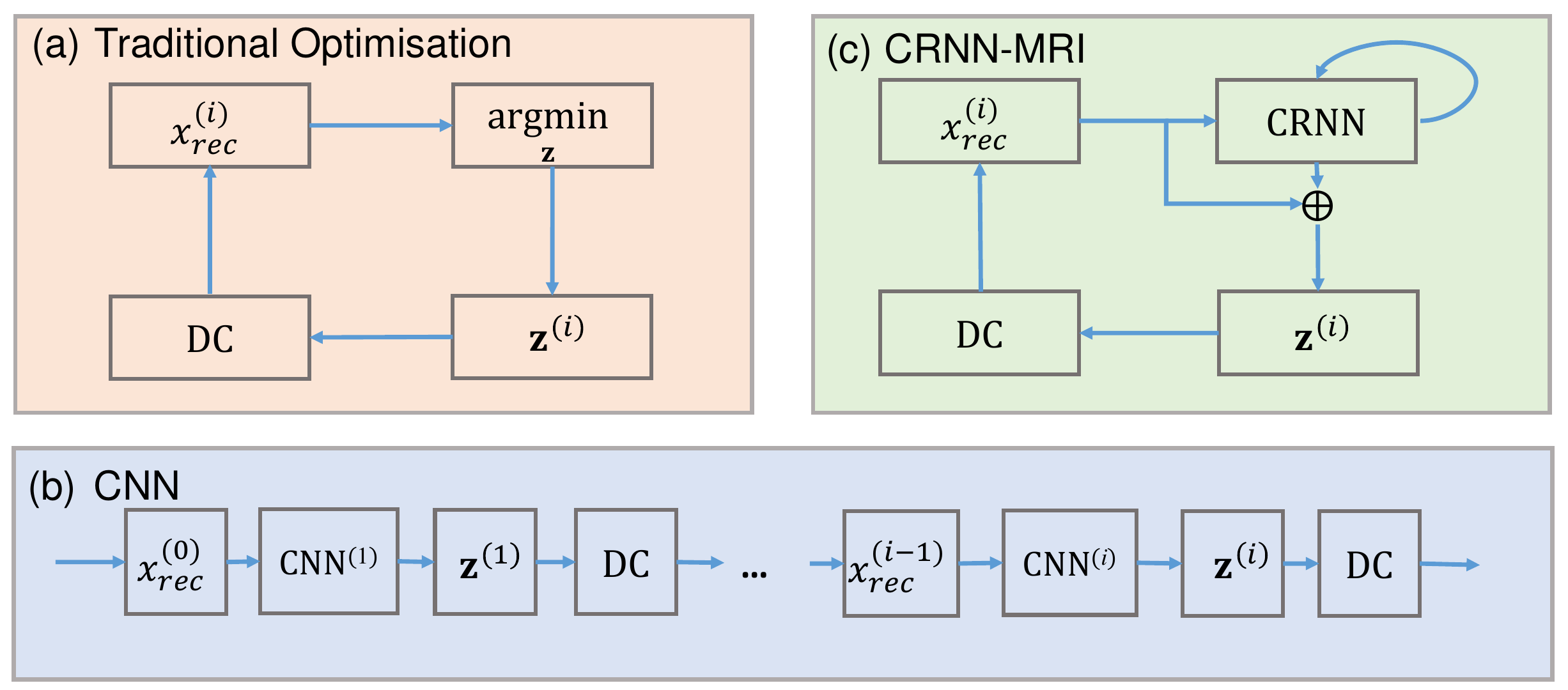}
\caption{(a) Traditional optimisation algorithm using variable splitting and alternate minimisation approach, (b) the optimisation unrolled into a deep convolutional network incorporating the data consistency step, and (c) the proposed architecture which models optimisation recurrence.}
\label{fig:opt}
\end{figure*}

Previous deep learning approaches such as Deep-ADMM net \cite{NIPS2016_6406} and method proposed by Schlemper et al. \cite{schlemper2017deep} unroll the traditional optimisation algorithm. Hence, their models learn a sequence of transition $\mbx^{(0)} \rightarrow \mbz^{(1)} \rightarrow \mbx^{(1)} \rightarrow \dots \rightarrow \mbz^{(N)} \rightarrow \mbx^{(N)}$ to reconstruct the image, where each state transition at stage $(i)$ is an operation such as convolutions independently parameterised by $\boldsymbol{\theta}$, nonlinearities or a data consistency step. { However, since the network implicitly learns some form of proximal operator at each iteration, it may be redundant to individually parameterise each step.} In our formulation, we model each optimisation stage $(i)$ as a learnt, \emph{recurrent}, forward encoding step $f_i(\mbx^{(i-1)},\mbz^{(i-1)};\boldsymbol{\theta}, \mby, \lambda, \Omega)$. { The difference is that now we use one model which performs proximal operator, however, it also allows itself to propagate information across iteration, making it adaptable for the changes across the optimisation steps. The detail will be discussed in the following section. }
The different strategies are illustrated in Fig \ref{fig:opt}. 

\subsection{CRNN for MRI reconstruction}

\begin{figure*}[!t]
\centering
\includegraphics[width=0.85\linewidth]{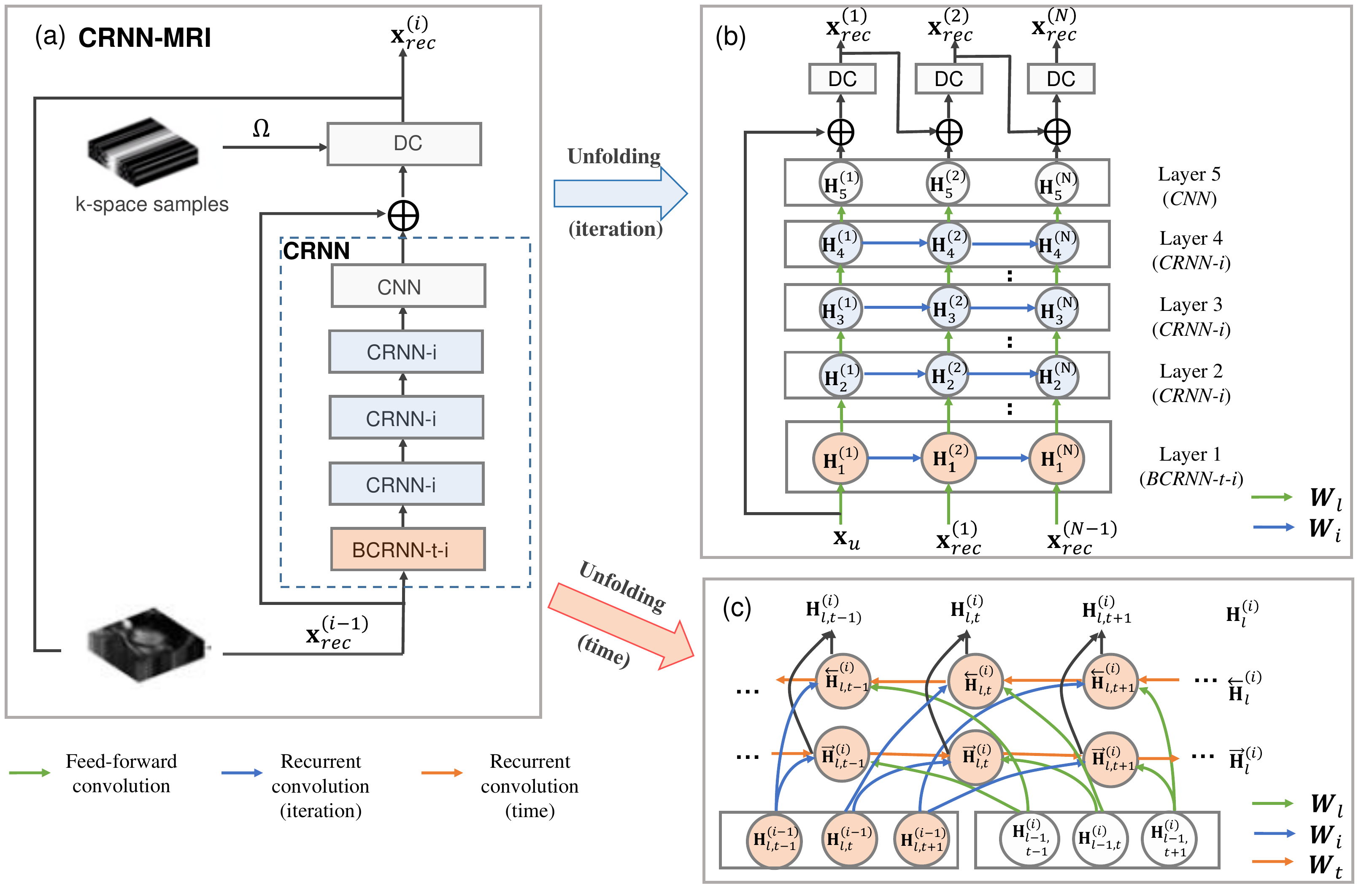}
\caption{(a) The overall architecture of proposed CRNN-MRI network for MRI reconstruction. (b) The structure of the proposed network when unfolded over iterations, in which ${\bf{x}}_{rec}^{(0)} = {{\bf{x}}_u}$. (c) The structure of BCRNN-t-i layer when unfolded over the time sequence. The green arrows indicate feed-forward convolutions {which are denoted by $W_l$}. The blue arrows ({$W_i$}) and red arrows ({$W_t$}) indicate recurrent convolutions over iterations and the time sequence respectively. {For simplicity, we use a single notation to denote weights for these convolutions at different layers. However, in the implementation, the weights are independent across layers.} }
\label{fig:RNN-fig}
\end{figure*}

RNN is a class of neural networks that makes use of sequential information to process sequences of inputs. They maintain an internal state of the network acting as a "memory", which allows RNNs to naturally lend themselves to the processing of sequential data. Inspired by iterative optimisation schemes of Eq. \ref{eq:alternate_minimisation}, we propose a novel convolutional RNN (CRNN) network. In the most general scope, our neural encoding model is defined as follows, 

\begin{equation}
{{\bf{x}}_{rec}} = {f_N}({f_{N - 1}}(\cdots({f_1}({\bf{x}}_{u})))),
\end{equation}
in which ${\bf{x}}_{rec}$ denotes the prediction of the network, ${\bf{x}}_{u}$ is the sequence of undersampled images with length $T$ and also the input of the network, $f_i({{\bf{x}}_u};\boldsymbol{\theta},\lambda ,\Omega )$ is the network function for each iteration of optimisation step, and $N$ is the number of iterations. We can compactly represent a single iteration $f_i$ of our network as follows:

\begin{subequations}
\begin{align}
{\bf{x}}_{rnn}^{(i)} &= {\bf{x}}_{rec}^{(i - 1)} + \textnormal{CRNN}({\bf{x}}_{rec}^{(i - 1)}), 
\\
{\bf{x}}_{rec}^{(i)} &= \textnormal{DC}({\bf{x}}_{rnn}^{(i)}; \mby, \lambda_0, \Omega), 
\end{align}
\end{subequations} 
where CRNN is a learnable block explained hereafter, DC is the data consistency step treated as a network layer,  ${\bf{x}}_{rec}^{(i)}$ is the progressive reconstruction of the undersampled image ${{\bf{x}}_u}$ at iteration $i$ with ${\bf{x}}_{rec}^{(0)} = {{\bf{x}}_u}$, ${\bf{x}}_{rnn}^{(i)}$ is the intermediate reconstruction image before the DC layer, and ${\bf{y}}$ is the acquired k-space samples. Note that the variables $\mbx_{rec}, \mbx_{rnn}$ are analogous to $\mbx, \mbz$ in Eq. \ref{eq:alternate_minimisation} respectively. Here, we use CRNN to encode the update step, which can be seen as one step of a gradient descent in the sense of objective minimisation, or a more general approximation function regressing the difference $\mbz^{(i+1)} - \mbx^{(i)}$, i.e. the distance required to move to the next state. Moreover, note that in every iteration, CRNN updates its internal state $\mathcal{H}$ given an input which is discussed shortly. As such, CRNN also allows information to be propagated efficiently across iterations, in contrast to the sequential models using CNNs which collapse the intermediate feature representation to ${\bf{z}}^{(i)}$.  

In order to exploit the dynamic nature and the temporal redundancy of our data, we further propose to jointly model the recurrence evolving over time for dynamic MRI reconstruction. The proposed CRNN-MRI network and CRNN block are shown in Fig. \ref{fig:RNN-fig}(a), in which CRNN block comprised of 5 components: 
\begin{enumerate}
\item bidirectional convolutional recurrent units evolving over time and iterations (BCRNN-t-i), 
\item convolutional recurrent units evolving over iterations only (CRNN-i), 
\item 2D convolutional neural network (CNN),  
\item residual connection and 
\item DC layers. 
\end{enumerate}
We introduce details of the components of our network in the following subsections.

\subsubsection{CRNN-i}
\label{CRNN-i}
As aforementioned, we encapsulate the iterative optimisation procedures explicitly with RNNs. In the CRNN-i unit, the iteration step is viewed as the sequential step in the vanilla RNN. If the network is unfolded over the iteration dimension, the network can be illustrated as in Fig. \ref{fig:RNN-fig}(b), where information is propagated between iterations. Here we use $\bf{H}$ to denote the feature representation of our sequence of frames throughout the network. ${\bf{H}}_{l}^{(i)}$ denotes the representation at layer $l$ (subscript) and iteration step $i$ (superscript). Therefore, at iteration $(i)$, given the input {${\bf{H}}_{l - 1}^{(i)}$} and the previous iteration's hidden state ${\bf{H}}_{l}^{(i - 1)}$,  the hidden state  ${\bf{H}}_{l}^{(i)}$ at layer $l$ of a CRNN-i unit can be formulated as:
 \begin{equation}
 {\bf{H}}_{l}^{(i)} = \sigma ({{\bf{W}}_l}*{{\bf{H}}_{l - 1}^{(i)}} + {{\bf{W}}_i}*{{\bf{H}}_{l}^{(i - 1)}}+{\bf{B}}_{l}).
 \end{equation}
 
Here $*$ represents convolution operation, ${{\bf{W}}_l}$ and ${{\bf{W}}_i}$ represent the filters of input-to-hidden convolutions and hidden-to-hidden recurrent convolutions evolving over iterations respectively, and ${\bf{B}}_{l}$ represents a bias term. Here ${\bf{H}}_{l}^{(i)}$ is the representation of the whole $T$ sequence with shape ($batchsize, T , n_c, D_x, D_y$), where $n_c$ is the number of channels which is 2 at the input and output but is greater while processing inside the network, and the convolutions are computed on the last two dimensions. The latent features are activated by the rectifier linear unit (ReLU) as a choice of nonlinearity, i.e. $\sigma(x)=max(0,x)$.

The CRNN-i unit offers several advantages compared to independently unrolling convolutional filters at each stage. Firstly, compared to CNNs where the latent representation from the previous state is not propagated, the hidden-to-hidden iteration connections in CRNN-i units allow contextual spatial information gathered at previous iterations to be passed to the future iterations. This enables the
reconstruction step at each iteration to be optimised not only based on the output image but also based on the hidden features from previous iterations, where the hidden connection
convolutions can "memorise" the useful features to avoid redundant computation.
Secondly, as the iteration number increases, the effective receptive field of a
CRNN-i unit in the spatial domain also expands whereas CNN resets it at each iteration. This property allows the network to further improve the reconstruction by allowing it to have better contextual support. In addition, since the weight parameters are shared across iterations, it greatly reduces the number of parameters compared to CNNs, potentially offering better generalization properties.

In this work, we use a vanilla RNN {\cite{elman1990finding}} to model the recurrence due to its simplicity.
Note this can be naturally generalised to other RNN units, such as long
short-term memory (LSTM) and gated recurrent unit (GRU), which are considered to
have better memory properties, although using these units would significantly increase computational complexity. 

\subsubsection{BCRNN-t-i}

Dynamic MR images exhibit high temporal redundancy, which is often exploited as a-priori knowledge to regularise the reconstruction. Hence, it is also beneficial for the network to learn the dynamics of sequences. To this extent, we propose a bidirectional convolutional recurrent unit (BCRNN-t-i) to exploit both temporal \emph{and} iteration dependencies jointly. BCRNN-t-i includes three convolution layers: one on the input which comes into the unit from the previous layer {indicated by the green arrows in Fig. \ref{fig:RNN-fig}(c)}, one on the hidden state from the past and future time frames {as shown by the red arrows}, and the one on the hidden state from the previous iteration of the unit ({blue arrows in Fig. \ref{fig:RNN-fig}(c)}). Note that we simultaneously consider temporal dependencies from past and future time frames, and the encoding weights are shared for both directions.  The output for the BCRNN-t-i layer is obtained by summing the feature maps learned from both directions. The illustration figure of the unit when it is unfolded over time sequence is shown in Fig. \ref{fig:RNN-fig}(c).

As we need to propagate information along temporal dimensions in this unit, here we introduce an additional index $t$ in the notation to represent the variables related with time frame $t$. Here ${\bf{H}}_{l,t}^{(i)}$ represents feature representations at $l$-th layer, time frame $t$, and at iteration $i$,  $\overrightarrow{\bf{H}}_{l,t}^{(i)}$ denotes the representations calculated when information is propagated forward inside the BCRNN-t-i unit, and similarly,  $\overleftarrow{\bf{H}}_{l,t}^{(i)}$ denotes the one in the backward direction. Therefore, for the formulation of BCRNN-t-i unit, given (1) the current input representation of the $l$-th layer at time frame $t$ and iteration step $i$, which is the output representation from  ($l-1$)-th layer  ${\bf{H}}_{l-1,t}^{(i)}$, (2) the previous iteration's hidden representation within the same layer ${{\bf{H}}_{l,t}^{(i-1)}}$, (3) the hidden representation of the past time frame $\overrightarrow{\bf{H}}_{l,t - 1}^{(i)}$, and the hidden representation of the future time frame $\overleftarrow{\bf{H}}_{l,t + 1}^{(i)}$, then the hidden state representation of the current $l$-th layer of time frame $t$ at iteration $i$, ${\bf{H}}_{l,t}^{(i)}$ with shape $(batchsize, n_c, D_x, D_y)$, can be formulated as: 
\begin{sequation}
\begin{aligned}
\overrightarrow{\bf{H}}_{l,t}^{(i)} &= \sigma ({{\bf{W}}_l}*{\bf{H}}_{l - 1,t}^{(i)} + {{\bf{W}}_t}*{\overrightarrow{\bf{H}}_{l,t-1}^{i}} + {{\bf{W}}_i}*{{\bf{H}}_{l,t}^{(i-1)}}+{\overrightarrow{\bf{B}}_{l}}),\\
\overleftarrow{\bf{H}}_{l,t}^{(i)} & = \sigma ({{\bf{W}}_l}*{{\bf{H}}_{l - 1,t}^{(i)}} + {{\bf{W}}_t}* {\overleftarrow{\bf{H}}_{l,t+1}^{(i)}} + {{\bf{W}}_i}*{{\bf{H}}_{l,t}^{(i-1)}}+{\overleftarrow{\bf{B}}_{l}}),\\
{\bf{H}}^{(i)}_{l,t} &= {\overrightarrow{\bf{H}}_{l,t}^{(i)}} + {\overleftarrow{\bf{H}}_{l,t}^{(i)}},
\end{aligned}
\end{sequation}
Similar to the notation in Section \ref{CRNN-i}, ${{\bf{W}}_t}$ represents the filters of recurrent convolutions evolving over time.  When $l=1$ and $i=1$, ${\bf{H}}_{0,t}^{(1)}={\bf{x}}_{{u}_{t}}$, that is the $t$-th frame of undersampled input data, and when $l=1$ and $i=2,...T$, ${\bf{H}}_{0,t}^{(i)}={\bf{x}}_{{rec}_{t}}^{(i-1)}$, which stands for the $t$-th frame of the intermediate reconstruction result from iteration $i-1$. For ${{\bf{H}}_{l,t}^{(0)}}$, $\overrightarrow{\bf{H}}_{l,0}^{(i)}$ and $\overleftarrow{\bf{H}}_{l,T+1}^{(i)}$, they are set to be zero initial hidden states. 

The temporal connections of BCRNN-t-i allow information to be propagated across the whole $T$
time frames, enabling it to learn the differences and
correlations of successive frames. The filter responses of recurrent
convolutions evolving over time express dynamic changing biases, which focus
on modelling the temporal changes across frames, while the filter responses of
recurrent convolutions over iterations focus on learning the spatial refinement
across consecutive iteration steps. In addition, we note that learning recurrent layers along the temporal
direction is different to using 3D convolution along the space and temporal direction. 3D convolution seeks
invariant features across space-time, hence several layers of 3D convolutions are
required before the information from the whole sequence can be propagated to a particular time frame. On the other hand, learning
recurrent 2D convolutions enables the model to easily and efficiently propagate the information
through time, which also yields fewer parameters and a lower computational cost.

In summary, the set of hidden states for a CRNN block to update at iteration $i$ is $\mathcal{H} = \{ {\bf{H}}_{l}^{(i)},  {\bf{H}}_{l,t}^{(i)}, {\overleftarrow{\bf{H}}_{l,t}^{(i)}}, {\overrightarrow{\bf{H}}_{l,t}^{(i)}} \}$, for $l = 1, \dots, L$ and $t = 1, \dots, T$, where $L$ is the total number of layers in the CRNN block and $T$ is the total number of time frames.

\subsection{Network Learning}

Given the training data $S$ of input-target pairs ${\left( {{{\bf{x}}_u},{{\bf{x}}_t}} \right)}$, the network learning proceeds by minimizing the pixel-wise mean squared error (MSE) between the predicted reconstructed MR image and the fully sampled ground truth data:
\begin{equation}
\mathcal{L}\left( \boldsymbol{\theta}  \right){\rm{ = }}\frac{1}{n_S}\sum\limits_{\left( {{{\bf{x}}_u},{{\bf{x}}_t}} \right) \in S} {\left\| {{{\bf{x}}_t} - {{\bf{x}}_{rec}}} \right\|_2^2}
\end{equation}
where $\boldsymbol{\theta} = \{ {\bf{W}}_l, {\bf{W}}_i, {\bf{W}}_t, {\bf{B}}_l \}$,  $l=1\dots L$, and ${n_S}$ stands for the number of samples in the training set $S$. Note that the total number of time sequences $T$ and iteration steps $N$ assumed by the network before performing the reconstruction is a free parameter that must be specified in advance. The network weights were initialised using He initialization \cite{he2015delving} and it was trained using the Adam optimiser \cite{kingma2014adam}. During training, gradients were hard-clipped to the range of $[-5,5]$ to mitigate the gradient explosion problem. The network was implemented in Python using Theano and Lasagne libraries.

\section{Experiments}
\subsection{Dataset and Implementation Details}

The proposed method was evaluated using a complex-valued MR dataset consisting of 10 fully sampled short-axis cardiac cine MR scans. Each scan contains a single slice SSFP acquisition with 30 temporal frames, which have a $320\times320$ mm field of view and 10 mm thickness. The raw data consists of 32-channel data with sampling matrix size $192\times 190$, which was then zero-filled to the matrix size $256\times256$. The raw multi-coil data was reconstructed using SENSE \cite{pruessmann1999sense} with no undersampling and retrospective gating. Coil sensitivity maps were normalized to a body coil image and used to produce a single complex-valued reconstructed image. In experiments, the complex valued images were back-transformed to regenerate k-space samples, simulating a fully sampled single-coil acquisition. The input undersampled image sequences were generated by randomly undersampling the k-space samples using Cartesian undersampling masks, with undersampling patterns adopted from \cite{jung2007improved}: for each frame the eight lowest spatial frequencies were acquired, and the sampling probability of $k$-space lines along the phase-encoding direction was determined by a zero-mean Gaussian distribution. Note that the undersampling rates are stated with respect to the matrix size of raw data, which is $192\times 190$.

The architecture of the proposed network used in the experiment is shown in Fig. \ref{fig:RNN-fig}: each iteration of the CRNN block contains five units: one layer of BCRNN-t-i, followed by three layers of CRNN-i units, and followed by a CNN unit. For all CRNN-i and BCRNN-t-i units, we used a kernel size $k=3$ and the number of filters was set to $n_{f}=64$ for Proposed-A and $n_{f}=128$ for Proposed-B in Table \ref{psnr}. The CNN after the CRNN-i units contains one convolution layer with $k=3$ and $n_{f}=2$, which projects the extracted representation back to the image domain which contains complex-valued images expressed using two channels. {For all convolutional layers, we used stride $=1$ and paddings with half the filter size (rounded down) on both size.} The output of the CRNN block is connected to the residual connection, which sums the output of the block with its input. Finally, we used DC layers on top of the CRNN output layers. During training, the iteration step is set to be $N=10$, and the time sequence for training is $T=30$. Note that this architecture is by no means optimal and more layers can be added to increase the ability of our network to better capture the data structures {(see Section \ref{variations of architecture} for comparisons)}. 

The evaluation was done via a 3-fold cross validation, {where for two folds we train on 7 subjects then test on 3 subjects, and for the remaining fold we train on 6 subjects and test on 4 subjects. While the original sequence has size $256 \times 256 \times T$, For the training, we extract patches of size $256 \times D_{patch} \times T$, where {$D_{patch}=32$} is the patch size and the direction of patch extraction corresponds to the frequency-encoding direction. Note that since we only consider Cartesian undersampling, the aliasing occurs only along the phase encoding direction, so patch extraction does not alter the aliasing artefact. Patch extraction as well as data augmentation was performed on-the-fly, with random affine and elastic transformations on the image data. Undersampling masks were also generated randomly following patterns in \cite{jung2007improved} for each input. During test time, the network trained on patches is directly applied on the whole sequence of the original image.} The minibatch size during the training was set to 1, and we observed that the performance can reach a plateau within $6 \times 10^4$ backpropagations.

\subsection{Evaluation Method}

We compared the proposed method with the representative algorithms of the CS-based dynamic MRI reconstruction, such as k-t FOCUSS \cite{jung2007improved} and k-t SLR \cite{lingala2011ktslr}, and two variants of 3D CNN networks named 3D CNN-S and 3D CNN in our experiments. The built baseline 3D CNN networks share the same architecture with the proposed CRNN-MRI network but all the recurrent units and 2D CNN units were replaced with 3D convolutional units, that is, in each iteration, the 3D CNN block contain 5 layers of 3D convolutions, one DC layer and a residual connection. Here 3D CNN-S refers to network sharing weights across iterations, however, this does not employ the hidden-to-hidden connection as in the CRNN-i unit. The 3D CNN-S architecture was chosen so as to make a fair comparison with the proposed model using a comparable number of network parameters. In contrast, 3D CNN refers to the network without weight sharing, in which the network capacity is $N=10$ times of that of 3D CNN-S, and approximately 12 times more than that of our first proposed method (Proposed-A). {For the 3D CNN approaches, the receptive field size is $11 \times 11 \times 11$, as the receptive field size is ``reset'' after each data consistency layer. In contrast, for the proposed method, due to the hidden connections between iterations and bidirectional temporal connections, by tracing the longest path of the convolution layers involved in the forward pass, including both temporal and iterative directions, in theory, the receptive field size is $309 \times 309 \times 30$ (154 layers of CNNs for the middle frame in a sequence of 30 frames). However, the network still may predominantly relies on local features coming from the partial reconstruction. Nevertheless, the RNN has the ability to exploit the features with larger filter size if needed, which is not the case for 3D CNNs.}

Reconstruction results were evaluated based on the following quantitative metrics: MSE, peak-to-noise-ratio (PSNR), structural similarity index (SSIM) \cite{wang2004image} and high frequency error norm (HFEN) \cite{ravishankar2011mr}. The choice of the these metrics was made to evaluate the reconstruction results with complimentary emphasis. MSE and PSNR were chosen to evaluate the overall accuracy of the reconstruction quality. SSIM put emphasis on image quality perception. HFEN was used to quantify the quality of the fine features and edges in the reconstructions, and here we employed the same filter specification as in \cite{ravishankar2011mr,miao2016accelerated} with the filter kernel size $15\times15$ pixels and a standard deviation of 1.5 pixels. For PSNR and SSIM, it is the higher the better, while for MSE and HFEN, it is the lower the better.

\begin{table*}[!t]
  \centering
  \caption{Performance comparisons (MSE, PSNR:dB, SSIM, and HFEN) on dynamic cardiac data with different acceleration rates. MSE is scaled to $10^{-3}$. The bold numbers are better results of the proposed methods than that of the other methods.}
  \label{psnr}
 
   \small
  \begin{tabular}{cccccccc}
  
    \toprule
  \multicolumn{2}{c}{Method} & {k-t FOCUSS} & {k-t SLR} & {3D CNN-S}   & 3D CNN & Proposed-A  & Proposed-B      \\
  \midrule
  \multicolumn{2}{c}{Capacity} & - & - &338,946  & 3,389,460 & \bf{262,020} & \bf{1,040,132}\\
  \midrule
    \multirow{4}{0.5cm}{$6\times$} & {MSE} & 0.592 (0.199) & 0.371(0.155) & 0.385 (0.124)  & {0.275 (0.096)} & \bf{0.261 (0.097)}&  \bf{0.201 (0.074)}\\
  & PSNR & 32.506 (1.516) & 34.632 (1.761) & 34.370 (1.526)  &{35.841 (1.470)} & \bf{36.096 (1.539)}&\bf{37.230 (1.559)} \\
    & SSIM & 0.953 (0.040) &	0.970 (0.033) &0.976 (0.008)   & {0.983 (0.005)} & \bf{0.985 (0.004)} &\bf{0.988 (0.003)} \\
  & HFEN & 0.211 (0.021) & 0.161 (0.016)& 0.170 (0.009) & {0.138 (0.013)}& \bf{0.131 (0.013)} &\bf{0.112 (0.010)} \\
    \midrule
    \multirow{4}{0.5cm}{$9\times$} & {MSE} & 1.234 (0.801) & 0.846 (0.572) & 0.929 (0.474)  &{0.605 (0.324)} & \bf{0.516 (0.255)}& \bf{0.405 (0.206)}\\
  & PSNR & 29.721 (2.339) & 31.409 (2.404) & 30.838 (2.246)  & {32.694 (2.179)} & \bf{33.281 (1.912)} & \bf{34.379 (2.017)}\\
    & SSIM &  0.922 (0.043) &	0.951 (0.025) 		  & 0.950 (0.016)&  {0.968 (0.010)}&\bf{0.972 (0.009)} & \bf{0.979 (0.007)}\\
  & HFEN & 0.310(0.041) &	0.260 (0.034) &0.280 (0.034)  & {0.215 (0.021)}&\bf{0.201 (0.025)} & \bf{0.173 (0.021)}\\
    \midrule
    \multirow{4}{0.5cm}{$11\times$} & {MSE} &1.909 (0.828) &	{1.237 (0.620)}& 1.472 (0.733)&  {0.742 (0.325)} & \bf{0.688 (0.290)} & \bf{0.610 (0.300)}\\

  & PSNR & 27.593 (2.038) &	{29.577 (2.211)} & 28.803 (2.151)  & {31.695 (1.985)} & \bf{31.986 (1.885)} & \bf{32.575 (1.987)}\\
    & SSIM &  	0.880 (0.060) &	0.924 (0.034) & {0.925 (0.022)}  & {0.960 (0.010)}&\bf {0.964 (0.009)} & \bf{0.968 (0.011)}\\
  & HFEN & 0.390 (0.023) &	0.327 (0.028) & 0.363 (0.041) &{0.257 (0.029)} & \bf{0.248 (0.033)} & \bf{0.227 (0.030)}\\
    \midrule
  \multicolumn{2}{c}{Time} & 15s & 451s & 8s  & 8s & \bf{3}s & \bf{6s}\\
    \bottomrule
  \end{tabular}
\end{table*}

\subsection{Results}

\begin{figure}[!t]
\centering
\includegraphics[width=.8\linewidth]{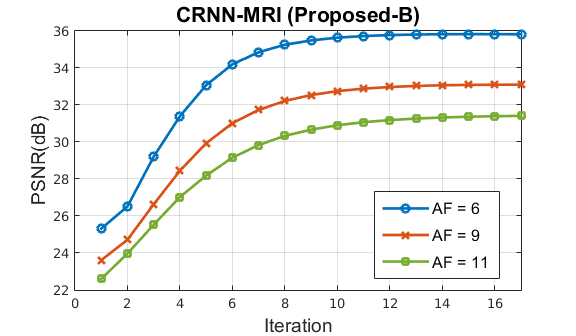}
\caption{Mean PSNR values (Proposed-B) vary with the number of iterations at test time on data with different acceleration factors. Here AF stands for acceleration factor.} 
\label{fig:iteration}
\end{figure}

\begin{figure*}[!t]
\centering
\includegraphics[width=1\linewidth]{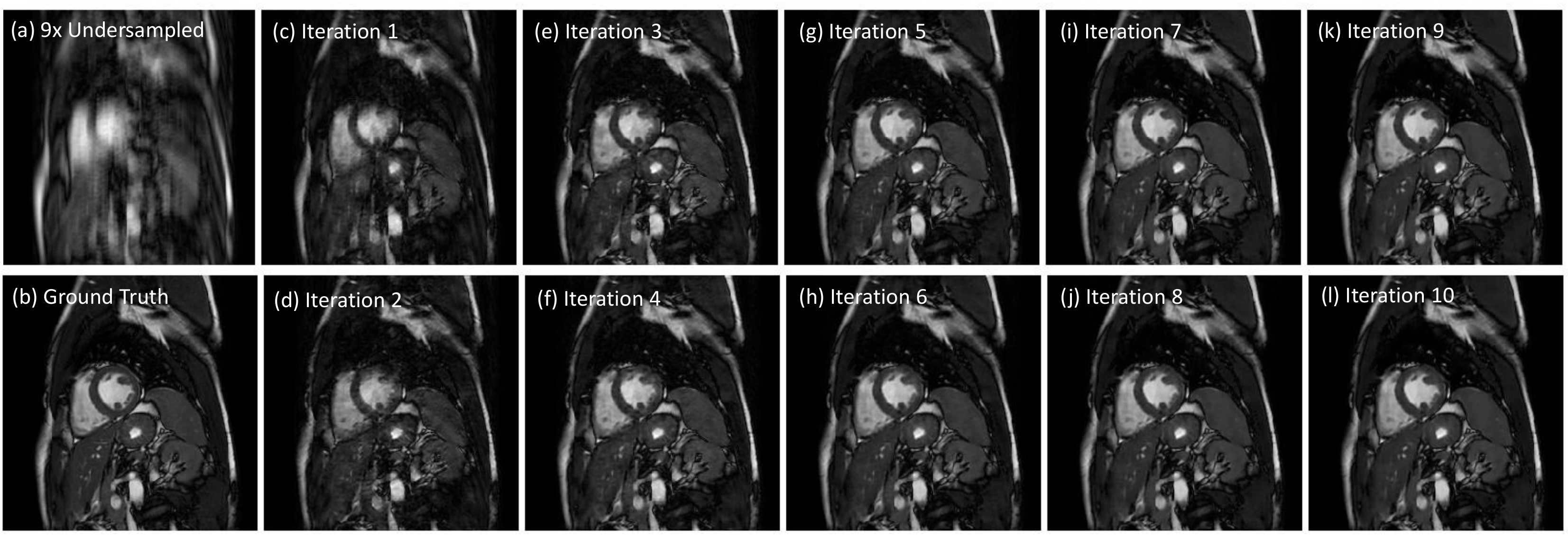}
\caption{Visualization results of intermediate steps during the iterations of a reconstruction. (a) Undersampled image by acceleration factor 9 (b) Ground Truth (c-l) Results from intermediate steps 1 to 10 in a reconstruction process.}
\label{fig:iterations}
\end{figure*}

The comparison results of all methods are reported in Table \ref{psnr}, where we evaluated the quantitative metrics, network capacity and reconstruction time. Numbers shown in Table \ref{psnr} are mean values of corresponding metrics with standard deviation of different subjects in parenthesis. Bold numbers in Table \ref{psnr} indicate the better performance of the proposed methods than the competing ones. Compared with the baseline method (k-t FOCUSS and k-t SLR), the proposed methods outperform them by a considerable margin at different acceleration rates. When compared with deep learning methods, note that the network capacity of Proposed-A is comparable with that of 3D CNN-S and the capacity of Propose-B is around one third of that of 3D CNN. Though their capacities are much smaller, both Proposed-A and Proposed-B outperform 3D CNN-S and 3D CNN for all acceleration rates by a large margin, which shows the competitiveness and effectiveness of our method. In addition, we can see a substantial improvement of the reconstruction results on all acceleration rates and in all metrics when the number of network parameters is increased for the proposed method (Proposed-B), and therefore we will only show the results from Proposed-B in the following. The number of iterations used by the network at test time is set to be the same as the training stage, which is $N=10$, however, if the iteration number is increased up to $N=17$, it shows an improvement of 0.324dB on average. Fig. \ref{fig:iteration} shows the model's performance varying with the number of iterations at test time. {Similarly, visualization results of intermediate steps during the iterations of a reconstruction from 9$\times$ undersampling data are shown in Fig. \ref{fig:iterations}, where we can observe the gradual improvement of the reconstruction quality from iteration step 1 to 10, which is consistent with the quantitative results as in Fig. \ref{fig:iteration}. } 

\begin{figure*}[!t]
\centering
\includegraphics[width=1\linewidth]{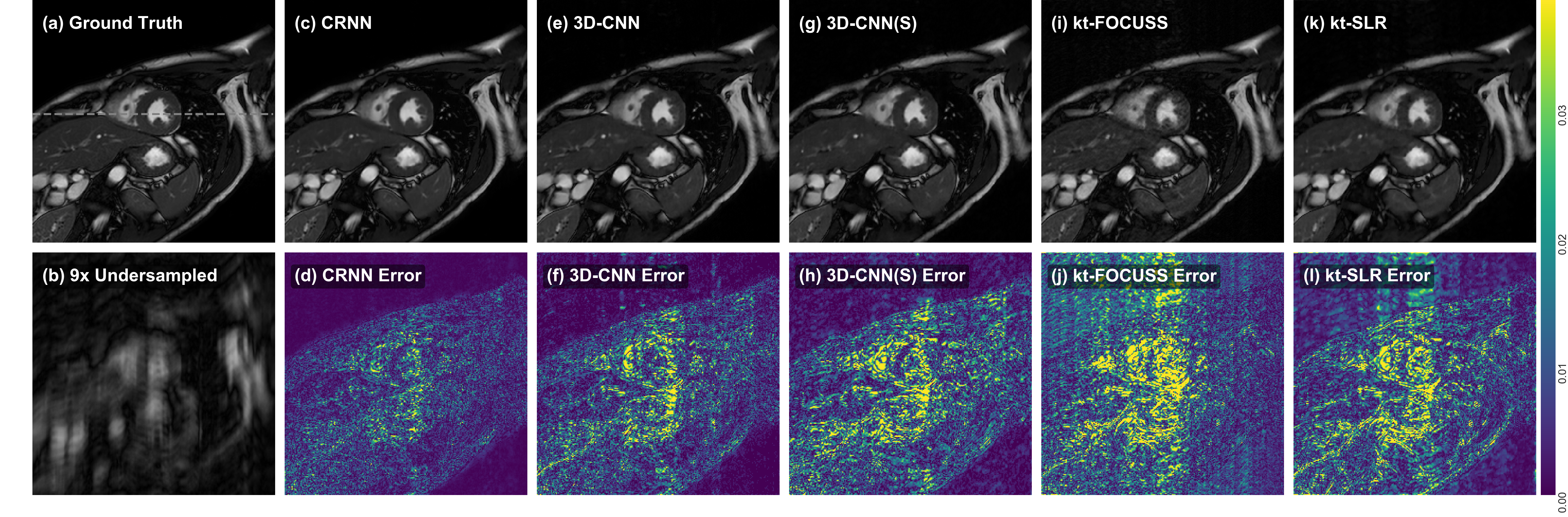}
\caption{The comparison of reconstructions on spatial dimension with their error maps. (a) Ground Truth (b) Undersampled image by acceleration factor 9 (c,d) Proposed-B (e,f) 3D CNN (g,h) 3D CNN-S (i,j) k-t FOCUSS (k,l) k-t SLR}
\label{fig:experiment1}
\end{figure*}

\begin{figure}[!t]
\centering
\includegraphics[width=1\linewidth]{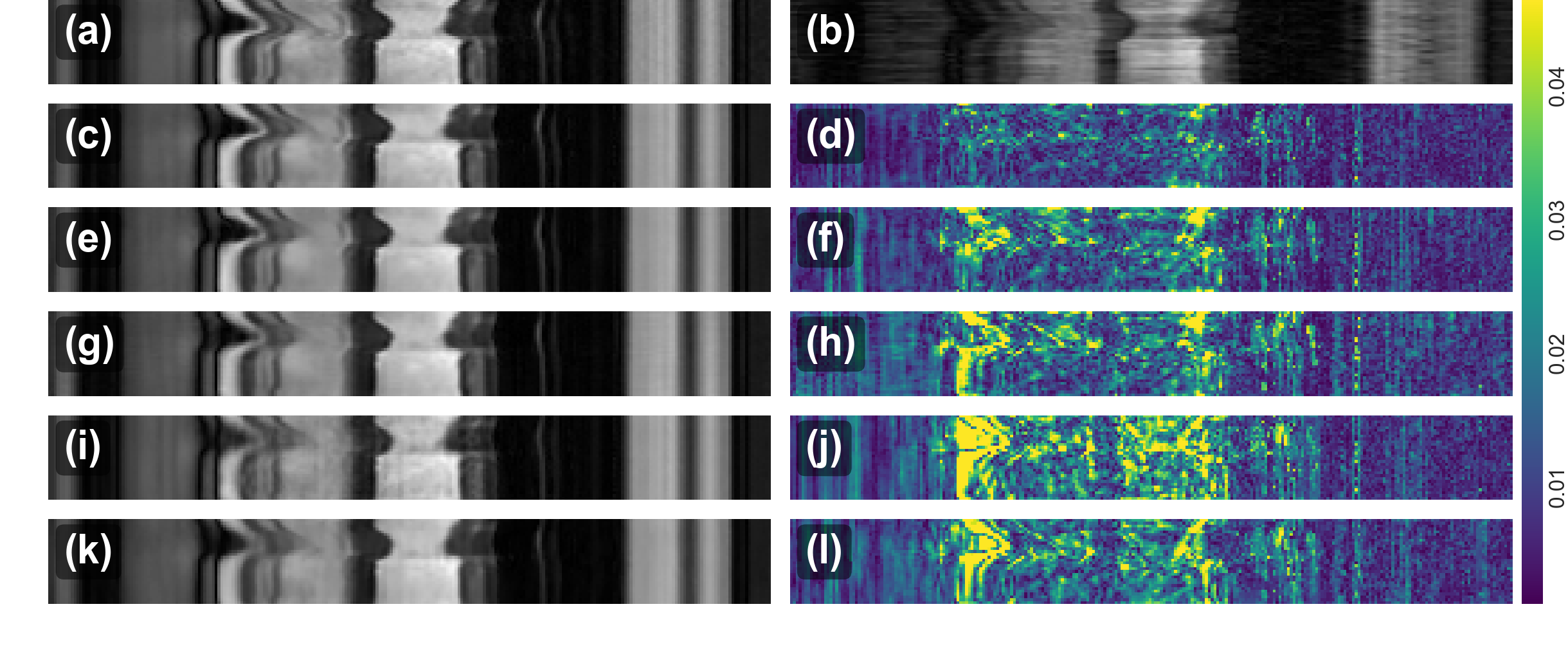}
\caption{The comparison of reconstructions along temporal dimension with their error maps. (a) Ground Truth (b) Undersampled image by acceleration factor 9 (c,d) Proposed-B (e,f) 3D CNN (g,h) 3D CNN-S (i,j) k-t FOCUSS (k,l) k-t SLR}
\label{fig:experiment2}
\end{figure}

A comparison of the visualization results of a reconstruction from $9\times$ acceleration is shown in Fig. \ref{fig:experiment1} with the reconstructed images and their corresponding error maps from different reconstruction methods. As one can see, our proposed model (Proposed-B) can produce more faithful reconstructions for those parts of the image around the myocardium where there are large temporal changes. This is reflected by the fact that RNNs effectively use a larger receptive field to capture the characteristics of aliasing seen within the anatomy. 
Their temporal profiles at $x=120$ are shown in Fig. \ref{fig:experiment2}. Similarly, one can see that the proposed model has overall much smaller error, faithfully modelling the dynamic data. It could be due to the fact that spatial and temporal features are learned separately in the proposed model while 3D CNN seeks invariant feature learning across space and time.

In terms of speed, the proposed RNN-based reconstruction is faster than the 3D CNN approaches because it only performs convolution along time once per iteration, removing the redundant 3D convolutions which are computationally expensive. Reconstruction time of 3D CNN and the proposed methods reported in Table \ref{psnr} were calculated on a GPU GeForce GTX 1080, and the time for k-t FOCUSS and k-t SLR were calculated on CPU.

{
\subsection{Variations of Architecture}
\label{variations of architecture}
In this section we show additional experiments to investigate the variants of the proposed architecture. First, we study the effects of recurrence over iteration and time, separately and jointly. In this study, we performed experiments on data set with undersampling factor 9, and the number of iterations was set to be 2 in order to simplify and speed up the training. Results are shown in Table \ref{ablation_study_1}, where we present the mean PSNR value via 3-fold cross validation. To isolate the effects of both recurrence in the module, we proposed to remove one of the recurrence each time. By removing the recurrence over time, the network architecture degrades to 4 CRNN-i + CNN layers, and it doesn't exploit temporal information in this case. If the recurrence over iterations is removed, the network architecture then becomes BCRNN-t + 4 CNN layers, without any hidden connections between iterations. Note that in all architectures, the last CNN layer only has 2 filters, which is used to simply aggregate the latent representation back to image space. Therefore, we employ a simple convolution layer for this. From Table \ref{ablation_study_1}, it can be observed that by removing any of the recurrent connections, the performance becomes worse compared with the proposed architecture with both recurrence jointly. This indicate that both of these recurrence contribute to the learning of the reconstruction. In particular, it is also been observed that by removing the temporal recurrence, the network's performance degrades greatly compared with the one removing the iteration recurrence. This can be explained that by removing the temporal recurrence, the problem degrades to a single frame reconstruction, while dynamic reconstruction has been proven to be much better than single frame reconstruction as there exists great temporal redundancies that can be exploited between frames.

\begin{table}[!t]
\centering
\caption{Performance comparisons on investigating the effects of each recurrence in the module. Reported results are the mean PSNR on data with undersampling factor 9 via 3-fold cross-validation. For this study, the number of iteration was set as 2.}
\label{ablation_study_1}
\begin{tabular}{cc}
\toprule
Architectures & PSNR (dB) \\
\midrule
4 CRNN-i + CNN (only iteration) & 21.41\\
BCRNN-t + 4 CNN (only temporal) & 26.62\\
BCRNN-t-i + 3 CRNN-i + CNN (Proposed) & 27.98\\
\bottomrule
\end{tabular}
\end{table}

In addition, we performed experiments on some other variants of the architecture, in particular, 4 layers of BCRNN-t-i with one layer of CNN, which has the highest capacity amongst all different combinations. Here we set the number of iterations to be 10. It can be observed that by incorporating temporal recurrent connections over all layers does improve the results over Proposed-A due to the more information propagated between frames. However, such design also increases the computations and more significantly, time required for training the network. Considering the trade off between performance and training time as well as the hardware constraints, we chose the particular design proposed. We agree that there could be more versions of the architectures that can lead to better performance and our particular design is by no means optimal. However, here we mainly aim to validate our proposed idea of exploiting both temporal and iterative reconstruction information for the problem, and the proposed architecture is satisfactory to show this. 

\begin{table}[!t]
\centering
\caption{Performance comparisons with different model architectures. Reported results are the mean PSNR on data with undersampling factor 9 via 3-fold cross-validation. (FPT: forward pass time; BPT: backward pass time)}
\label{ablation_study_2}
\begin{tabular}{ccccc}
\toprule
Architectures & PSNR (dB) & FPT & BPT & Training Time\\
\midrule
4 BCRNN-t-i + CNN & 34.18 & 0.94s & 5.97s & 96h\\
Proposed-A & 33.28 & 0.45s & 1.39s & 38h\\
Proposed-B & 34.38 & 0.90s & 2.59s & 58h\\
\bottomrule
\end{tabular}
\end{table}

}

{
\subsection{Feature Map Analysis}

\begin{figure*}[!t]
\centering
\includegraphics[width=\linewidth]{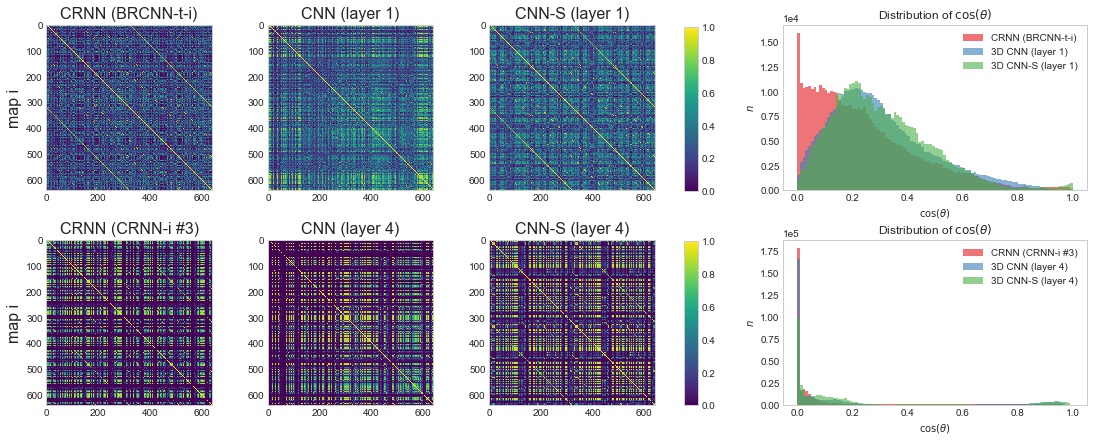}
\caption{Cosine distances for the feature maps extracted from $i$th-layer of the subnetworks across 10 cascades/iterations. Top row shows $i=1$, which corresponds to BRCNN-t-i unit for CRNN, 1st convolution layers for 3D-CNN and 3D-CNN-S. Bottom row shows $i=4$, which corresponds to the third CRNN-i unit for CRNN, 4th convolution layers for 3D-CNN and 3D-CNN-S. In general, the distribution of $\cos(\theta)$ is closer to 0 for CRNN than for the CNN's.}
\label{fig:feat_sim}
\end{figure*}

\begin{figure*}[!t]
\centering
\includegraphics[width=\linewidth]{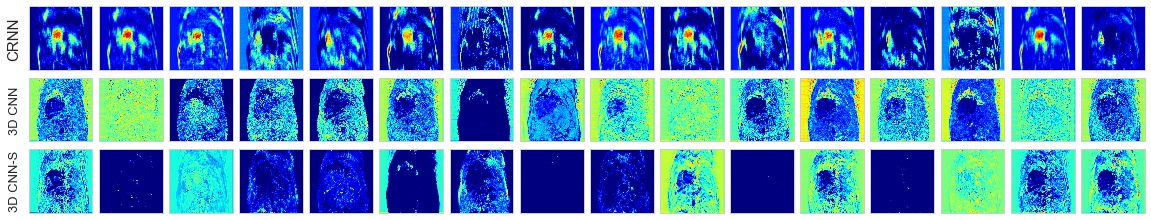}
\caption{Examples of the feature maps from the CRNN-MRI (Proposed-A), 3D CNN and 3D CNN-S, at iteration 10}
\label{fig:feat_maps}
\end{figure*}

In this section we study further whether the proposed architecture helps to obtain better feature representations. CRNN (Proposed-A), 3D-CNN and 3D-CNN-S all have the subnetworks composed of 5 units/layers with 64 channels for the first four, allowing us to directly compare the $i$-th layer of representations of the subnetworks for $i=1,\dots,4$. From one test subject, we extract the feature representations of the subnetwork across 10 cascades/iterations. By treating each channel as a separate feature map, we obtain 640 feature maps for each layer $i$ aggregated across iteration. We use the cosine distance $d(A,B)= A^T B /\|A\|\|B\| = \cos(\theta)$ to compute the similarity between these activation maps for $i \in \{1, 4\}$. If two feature maps are orthogonal, then $\cos(\theta)=0$ and if two feature maps are linearly correlated, then $\cos(\theta)=1$. Geometrically, this supports the interpretation that if the cosine distance is small for all the feature map pairs, then the network is likely to be capturing diverse patterns. The result is summarised in Fig. \ref{fig:feat_sim}, where the similarity measure is visualised as a matrix, as well as their distributions is plotted for each network. 

We can see that for both $i \in \{1, 4\}$, the layers from CRNN appears to have geometrically more orthogonal feature maps. One can also observe that in general, layer 1 has higher redundancy compared to layer 4. In particular, the diagonal yellow stripes can be observed for CNN-S and CRNN, due to parameter-sharing for each cascade. This is not observed in 3D-CNN, even though many features do have high similarity. In Fig. \ref{fig:feat_maps} we show examples of the feature maps from layer 4 (3rd CRNN-i for CRNN, 4th convolution layers for 3D-CNN and 3D-CNN-S) at iteration/cascade 10 {of each network during the forward pass. We selected 16 feature maps out of 64 by firstly clustering them into 16 groups, and then randomly chose one feature map from each group to show as representative feature maps in Fig. \ref{fig:feat_maps}. These feature maps show the activations learned from different networks and is colour-coded (blue corresponds to low activation whereas red corresponds to high activation)}. We see that CRNN's features look significantly different from CNN. In particular, one can observe that some are activated by the dynamic region, and some are particularly sensitive to regions around the left and/or right ventricle. 
}

\section{Discussion}

In this work, we have demonstrated that the presented network is capable of producing faithful image reconstructions from highly undersampled data, both in terms of various quantitative metrics as well as inspection of error maps. In contrast to unrolled deep network architectures proposed previously, we modelled the recurrent nature of the optimisation iteration using hidden representations with the ability to retain and propagate information across the optimisation steps. Compared with 3D CNN models, the proposed methods have a much lower network capacity but still have a higher accuracy, reflecting the effectiveness of our architecture. This is due to the ability of the proposed RNN units to increase the receptive field size while iteration steps increase, as well as to efficiently propagate information across the temporal direction. {In fact, for accelerated imaging, higher undersampling factors significantly add aliasing to the initial zero-filled reconstruction, making the reconstruction more challenging. This suggests that while the 3D CNN possesses higher modelling capacity owing to its large number of parameters, it may not necessarily be an ideal architecture to perform dynamic MR reconstruction, presumably because the simple CNN is not as efficient as propagating the information across the whole sequence. Besides, for the 3D CNN approaches, it is also observed that it is not able to denoise the background region. This could be explained by the fact that 3D CNN only exploits local information due to the small receptive field size it used, while in contrast, the proposed CRNN improves the denoising of the background region because of its larger receptive field sizes.} 

{Furthermore, when exploring the intermediate feature activations, we observed that the pair-wise cosine distances for CRNN were smaller than those for the 3D-CNNs.  We speculate that this is because CRNN has hidden connections across the iterations allowing it to propagate information better and make the end-to-end reconstruction process more dynamic, generating less redundant representations. On a contrary, 3D-CNNs needs to rebuild the feature maps at every iteration, which is likely to increase repetitive computations. In addition, qualitatively, the activation map of CRNN showed high sensitivity to anatomical regions/dynamic regions. This is likely due to the fact that CRNN has increased receptive field size as well as temporal units, allowing the network to recognise larger/dynamic objects better. In CNNs, one can also observe that there are features activated by the myocardial regions, however, the activation is more homogeneous across the image, due to smaller receptive field size. This hints that CRNN can better capture high level information.
}

In this work, we modeled the recurrence using the relatively simple (vanilla) RNN architecture. For the future work, we will explore other recurrent units such as LSTM or GRU. As they are trained to explicitly select what to remember, they may allow the units to better control the flow of information and could reduce the number of iterations required for the network to generate high-quality output. {Also, incorporating recurrent redundancy in k-space domain into the proposed CRNN-MRI network is likely to improve the result, and will form part of our future work.} In addition, we have found that the majority of errors between the reconstructed image and the fully sampled image lie at the part where motion exists, indicating that motion exhibits a challenge for such dynamic sequence reconstruction. Thus it will be interesting to explore {more efficient ways that can improve the reconstruction quality while faithfully preserving cardiac motion.}  Additionally, current analysis only considers a single coil setup. In the future, we will also aim at investigating such methods in a scenario where multiple coil data from parallel MR imaging can be used jointly for higher acceleration acquisition.

\section{Conclusion}

Inspired by variable splitting and alternate minimisation strategies, we have presented an end-to-end deep learning solution, CRNN-MRI, for accelerated dynamic MRI reconstruction, with a forward, CRNN block implicitly learning iterative denoising interleaved by data consistency layers to enforce data fidelity. In particular, the CRNN architecture is composed of the proposed novel variants of convolutional recurrent unit which evolves over two dimensions: time and iterations. The proposed network is able to learn both the temporal dependency and the iterative reconstruction process effectively, and outperformed the other competing methods in terms of both reconstruction accuracy and speed for different undersampling rates. 


\bibliographystyle{IEEEtran}
\bibliography{IEEEabrv,ref}

\end{document}